\documentclass[manuscript, nonacm]{acmart}

\usepackage{enumitem}
\usepackage{colortbl}
\usepackage{booktabs}
\usepackage{adjustbox}
\usepackage{multirow}
\usepackage{wrapfig}
\usepackage{eso-pic} 

\newcommand{\topaz}[0]{\textsc{Topaz}}

\AtBeginDocument{%
  }

\usepackage{tcolorbox}
\tcbuselibrary{skins, breakable}
\newtcolorbox{promptbox}[1][]{
  colback=gray!5, colframe=gray!60,
  fonttitle=\bfseries\small, title={#1},
  breakable, enhanced,
  left=6pt, right=6pt, top=4pt, bottom=4pt,
  boxrule=0.5pt, sharp corners,
}

\setcopyright{acmlicensed}
\copyrightyear{2026}
\acmYear{2026}
\acmDOI{XXXXXXX.XXXXXXX}
\acmConference[CHI '26]{}{June 03--05,
  2018}{Barcelona, Spain}
\begin{document}

\AddToShipoutPictureBG*{
  \AtPageUpperLeft{
    \put(0,-30){\makebox[\paperwidth][c]{In Proceedings of the 6th Human-Centered Explainable AI (HCXAI) workshop at CHI 2026, Barcelona (Spotlight)}}
  }
}

\title{Explainable Model Routing for Agentic Workflows}

\author{Mika Okamoto}
\orcid{0009-0001-4247-6635}
\email{mokamoto7@gatech.edu}

\author{Ansel Kaplan Erol}
\orcid{0009-0000-3149-075X}


\author{Mark Riedl}
\orcid{0000-0001-5283-6588}
\email{riedl@cc.gatech.edu}
\affiliation{%
  \institution{Georgia Institute of Technology}
  \city{Atlanta}
  \state{Georgia}
  \country{USA}
}

\renewcommand{\shortauthors}{Okamoto et al.}

\begin{abstract}
Modern agentic workflows decompose complex tasks into specialized subtasks and route them to diverse models to minimize cost without sacrificing quality. However, current routing architectures focus exclusively on performance optimization, leaving underlying trade-offs between model capability and cost unrecorded. Without clear rationale, developers cannot distinguish between \textit{intelligent efficiency}---using specialized models for appropriate tasks---and \textit{latent failures} caused by budget-driven model selection. 
We present \topaz, a framework that introduces formal auditability to agentic routing. \topaz~replaces silent model assignments with an inherently interpretable router that incorporates three components: (i) skill-based profiling that synthesizes performance across diverse benchmarks into granular capability profiles (ii) fully traceable routing algorithms that utilize budget-based and multi-objective optimization to produce clear traces of how skill-match scores were weighed against costs, and (iii) developer-facing explanations that translate these traces into natural language, allowing users to audit system logic and iteratively tune the cost-quality tradeoff. By making routing decisions interpretable, \topaz~enables users to understand, trust, and meaningfully steer routed agentic systems.
\end{abstract}



\received{19 February 2026}
\received[accepted]{18 March 2026}
\received[revised]{26 March 2026}

\maketitle

\section{Introduction}
As AI systems shift from monolithic models to composite agentic workflows, developers are increasingly employing \textit{model routing} to balance performance and cost across a system. By dynamically routing each input to the most suitable LLM within a diverse collection of models (e.g., routing simple queries to a cheaper model while reserving frontier models for complex reasoning), routing systems achieve significant efficiency gains~\citep{chen2024frugalgpt, yue2025masrouter, ong2024routellm, ding2024hybridllm}. 
Although a promising strategy for scaling workloads, routing also introduces novel explainability challenges, since developers now need to understand the criteria used to route queries between different LLM models.

Traditional interpretability explains why a model made a prediction.
Agentic routing, however, requires explaining why a sequence of models was selected in human-centered terms that developers can act on: whether task requirements were identified correctly across the workflow and whether budget constraints were met.
Current routing systems offer little support for this kind of reasoning, presenting model assignments as \textit{opaque decisions} with limited explanation or opportunities for stakeholder participation~\cite{graphrouter2025, chen2024frugalgpt, yue2025masrouter}.
Applying traditional post-hoc XAI techniques to those routing systems surfaces optimization internals---confidence thresholds or learned decision boundaries---rather than actionable reasoning about model-task fit.
Consequently, developers debugging pipelines struggle to diagnose failures and determine whether cost optimizations represent legitimate efficiency gains or critical quality compromises. 
Absent grounded explanations, developers must either blindly trust the routing system, manually audit every decision, or bypass routing entirely and rely on the most expensive frontier models---none of which scale.

Explainable routing poses three challenges. First, transparent capability profiling demands granular, skill-level signals, yet standard benchmarks reduce model performance to aggregate scores~\cite{wang2024mmlupro, chiang2024lmarena, evaltree2025}. Second, agent routing decisions arise from the interdependent interaction of task complexity, skill requirements, and cost, making individual criteria difficult to isolate and audit. Third, explanations of these decisions are prone to post-hoc rationalization that sounds plausible but fails to reflect actual decision logic, leaving developers unable to diagnose issues or improve their systems. 

To address these challenges, we present \topaz, an inherently interpretable framework for \textit{explainable routing in agentic settings}.
\topaz~comprises three stages: (1) \textit{Skill-based profiling} to decompose benchmarks, model capabilities, and task requirements into a shared skill taxonomy; (2) \textit{Cost-aware routing} via fixed-budget and multi-objective optimization to balance quality and cost; and (3) \textit{Developer-facing explanation generation} that synthesizes routing traces into natural language rationale, enabling developers to verify routing logic, and iteratively refine cost-quality preferences. \topaz~thus extends the frontier of explainability from the \textit{content} of single-model predictions to the \textit{context} of agentic routing, establishing a foundation for trustworthy agentic systems. In summary, our contributions are four-fold:

\begin{itemize}[nosep]
\item We highlight a critical deficit in agentic routing XAI---the lack of human-centered explainability for routing behavior---exposing key challenges and open questions in achieving transparent, effective agent routing.
\item We introduce a novel, domain-agnostic, and accessible approach for synthesizing public benchmarks into capability profiles, enabling transparent model analysis without excessive compute or data burdens.
\item We formulate two fully-traceable routing algorithms for assigning workflow tasks to models: one for planning under strict budgets and the other for general heuristic optimization, showcasing efficacy via case studies. 
\item We provide faithful and actionable insights based on intermediate computations from our routing algorithms, enabling developers to audit model assignments and iteratively tune their agent's cost-quality tradeoffs.

\end{itemize}

\begin{figure}
    \vspace{-1 ex}
    \centering
    \includegraphics[width=0.8\linewidth]{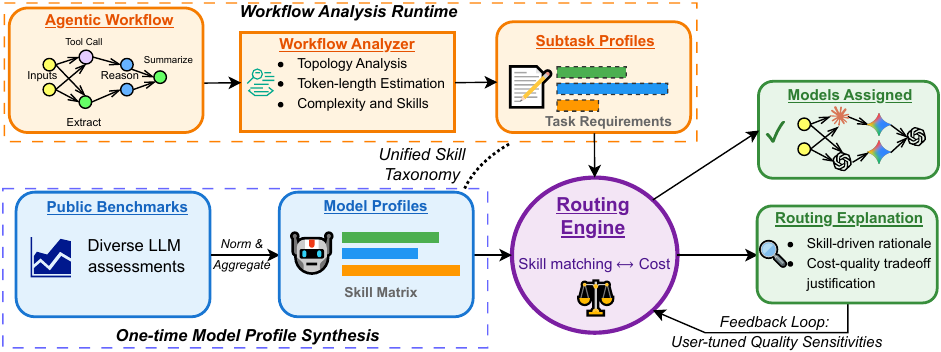}
    \Description{Architecture diagram showing two profiling pipelines feeding into a central routing engine. The bottom pipeline synthesizes public benchmarks into model capability profiles. The top pipeline analyzes an agentic workflow's subtasks for complexity, token-length, and skill requirements. The pipelines share a unified skill taxonomy. The routing engine balances skill matching against cost, producing model assignments and a routing explanation with skill-driven rationale and cost-quality tradeoff justification, with a feedback loop for user-tuned quality sensitivities.}
    \caption{\topaz~ Architecture. Public benchmarks are synthesized to form model capability profiles. Then, for a new agentic workflow, each subtask is analyzed for complexity and skill requirements. The \topaz~routing engine balances skill match and cost, yielding model assignments for each subtask while providing explainable traces for developers.}
    \label{fig:architecture}
    \vspace{-4 ex}
\end{figure}

\section{Related Work}

Cost-oriented routing has emerged as a practical response to the economic realities of LLM deployment. Cascade approaches escalate queries through increasingly expensive models until confidence thresholds are met~\citep{chen2024frugalgpt, aggarwal2024automix}, while learned routers predict query difficulty or preference-based quality to assign models directly~\citep{ding2024hybridllm, ong2024routellm}. \citet{dekoninck2025cascade} unifies these paradigms into a theoretically grounded framework, and Router-R1~\citep{zhang2025routerr1} extends routing to sequential multi-model coordination via reinforcement learning. These systems optimize cost-quality tradeoffs effectively but lack routing decision transparency, relying on opaque or latent mechanisms for evaluating quality or assigning models.


Successful routing requires understanding what a model is good at, not just how generally competent or cheap it is. FLASK~\citep{ye2024flask} evaluates models across fine-grained skill dimensions, exposing variance that aggregate scores mask, and Skill-Slices~\citep{moayeri2024skill} shows that skill-based routing improves accuracy. These approaches provide necessary granular capability assessments but do not explain routing. \textsc{BELLA}~\citep{beecatdaughter} extends this by grounding single-query routing in explainable skill decompositions, but does not address multi-task agentic workflows with interdependent decisions. 



The HCXAI community has emphasized that effective explanations must account for \textit{who} needs to understand \textit{what}~\citep{ehsan2021expanding, liao2024questioning}, a framing we adopt for orchestration decisions. \topaz~aims to provide effective explanations by combining local explanations, referring to why each task was routed to a particular model, and global rationale that characterizes broader, cross-task routing patterns and tradeoffs, a well-established technique in interpretable ML~\citep{ribeiro-etal-2016-trust}. Thus, \topaz~bridges XAI and routing: where prior XAI explains inference and prior routing optimizes selection, \topaz~is a novel router that makes orchestration decisions inherently explainable---why this model for this task at this cost.
\section{Design and Methods}
Explainable model routing requires (i) fine-grained capability profiling, (ii) decomposable cost-quality tradeoffs, and (iii) faithful and useful explanations.
\topaz~ is thus motivated by the research question: \textit{How can model routing decisions in agentic pipelines be grounded in human-interpretable quantities that support genuine understanding and not simply post-hoc justification?} We answer this question through designing a system that bases explanations on actual numerical traces used for routing decisions, balancing quality, inferred from skill alignment, with estimated cost.

\vspace{-0.7 em}
\subsection{Skill-Based Profiles for Understanding Models and Agentic Workflows}
\label{section:skill_profiles}
\topaz~routes agentic subtasks to LLMs by matching task requirements against model capabilities, both expressed in a shared, human-interpretable skill space. We define a skill set $\mathcal{S} = \{s_1, \ldots, s_k\}$ (e.g., logical reasoning, writing quality), where each skill has a natural-language description. To profile both benchmarks and tasks against $\mathcal{S}$, we prompt an LLM with a description and example input to obtain an $L_1$-normalized distribution of non-negative skill weights, enabling direct comparison between what models can do and what tasks require, grounded in interpretable skills. 

\textit{Synthesizing Benchmarks into Model Profiles.} We profile public benchmarks $b \in \mathcal{B}$ to obtain skill weights $w_{b,s}$, and collect third-party evaluation scores $S_{m, b}$ for each model $m \in M$. After 0-max normalizing scores as $\tilde{S}_{m,b} = \frac{S_{m,b}}{S^\text{max}_b}$ where $S^\text{max}_b$ is the best score on $b$ across $M$, we compute each model's per-skill capability score as:
\vspace{-1 ex}
\begin{equation}
    C_{m,s} = \frac{\sum_{b} \tilde{S}_{m,b} \cdot w_{b,s}}{\sum_{b} w_{b,s}},
    \label{eq:capability}
\end{equation}
where the denominator normalizes against the representation of skill $s$ across benchmarks $\mathcal{B}$. These profiles are \textit{static}, recomputed only when the model, benchmark, or skill pool changes.

\textit{Building Task Profiles for Agentic Workflows.} When a user submits an agentic workflow, they specify subtasks $t \in \mathcal{T}$ with descriptions, which \topaz~ profiles for skill requirements $R_{t,s}$. The LLM profiler also jointly analyzes the subtasks to extract task complexity $k_t$, estimated input and output token counts $\sigma_{\text{in}} \text{ and } \sigma_{\text{out}}$, and quality-sensitivity $q_t$ (how critical performance is for this subtask) for each task $t \in \mathcal{T}$. Quality sensitivity $q_t$ is also user-adjustable (Section \ref{sec:feedback_loop}).

\vspace{-0.9em}
\subsection{Cost Models for API-based LLM Inference}
\label{section:cost_modeling}
The absolute cost of routing a subtask $t$ to model $m$ is $\text{Cost}_{\text{abs}}(m, \sigma_t) = \sigma^{\text{in}}_tp_{m}^{\text{in}} + \sigma^{\text{out}}_tp_{m}^{\text{out}}$, where $p_m^{\text{in}}, p_m^{\text{out}}$ are per-token prices and $\sigma^{\text{in}}_t, \sigma^{\text{out}}_t$ are subtask token count estimates. However, accurately predicting response length before generation is unreliable~\citep{zheng2023responselengthperceptionsequence}, so we propose an alternate, relative pricing mechanism that requires only an estimate of the input/output skew $\sigma^{\text{i/o}}_t = \frac{\sigma^{\text{in}}_t}{\sigma^{\text{in}}_t + \sigma^{\text{out}}_t}$. Given $\sigma^{\text{i/o}}_t$, \hspace{2 pt}the relative price is 
%
%
$\text{Cost}_\text{rel}(m, \sigma^{\text{i/o}}_t) = \sigma^{\text{i/o}}_t \cdot p_m^{\text{in}}  +  (1 - \sigma^{\text{i/o}}_t) \cdot p_m^{\text{out}}$.
\quad Since this yields a per-token rate rather than an absolute cost, we min-max normalize $\text{Cost}_\text{rel}(m, \sigma^{\text{i/o}}_t)$ against the cheapest and most expensive models in model set $\mathcal{M}$ to obtain a comparable cost penalty $\text{Cost}^{\text{min-max}}_{\text{rel}}(m, \sigma^{\text{i/o}}_t)$.

\vspace{-0.9em}
\subsection{Routing Algorithms for Interpretable Task-to-Model Assignment}
\label{section:routing_algorithms}
\paragraph{Skill Match Score.}
We quantify the fit between capabilities and requirements, capping credit at satisfaction (exceeding requirements provides no benefit):
\vspace{-1.4 ex}
\begin{equation}
    \text{Match}_{m,t} = \sum_{s \in \mathcal{S}} \underbrace{\min\left(1, \frac{C_{m,s}}{k_t \cdot R_{t,s}}\right)}_{\text{Skill Fulfillment Ratio}} \cdot R_{t,s}
\end{equation}
This score represents the expected output quality from model $m$ on subtask $t$. We outline two routing algorithms: one that minimizes costs to quality constraints and one that maximizes quality subject to a budget.

\textit{Objective-based Routing.}
Each subtask $t \in \mathcal{T}$ is routed independently by optimizing a weighted trade-off between quality and cost to achieve a globally directed but locally adapted balance between cost and quality. Quality and cost weights are coupled through $q_t$ (local quality-sensitivity) and $c_{\text{global}}$ (global cost sensitivity), with a floor $\varepsilon = 0.01$ ensuring neither factor fully vanishes at extreme settings. For each task, we assign:
\vspace{-1 ex}
\begin{equation}
    m^* = \arg\max_{m \in \mathcal{M}} \left[
      q_t \cdot \max\!\big(1 - c_{\text{global}}, \varepsilon \big)
      \cdot \text{Match}_{m,t}
      \;-\;
      c_{\text{global}} \cdot\max\!\big(1 - q_t, \varepsilon \,\big)
      \cdot \text{Cost}^{\text{min-max}}_{\text{rel}}(m, \sigma^{\text{i/o}}_t)
    \right],
\end{equation}

\textit{Budget-based Routing.}
For a subtask sequence $(t_1, \ldots, t_n)$ with budget $B$, we maximize overall quality via dynamic programming. We find the best achievable quality $Q$ when assigning the $i^\mathrm{th}$ task with remaining budget $c$ as:
\begin{equation}
    Q[i, c] = \max_{m} \left(Q[i-1, c - \text{Cost}_{\text{abs}}(m, \sigma_t)] + q_{t_i}\text{Match}_{m,t_i} \right),
\end{equation}
where $q_{t_i}\text{Match}_{m,t_i}$ is quality and $\text{Cost}_{\text{abs}}(m, \sigma_t)$ is absolute cost. Model assignments are recovered via back-tracing.

\subsection{Explanation Generation}

\topaz~generates developer-facing explanations by synthesizing the numerical routing decisions into natural language summaries. The system maintains a structured explanation log that records: (1) user configuration, containing cost sensitivity $c_{\text{global}}$, quality sensitivity $q_t$, subtask specifications; (2) intermediate calculations, consisting of skill match scores $\text{Match}_{m,t}$ and cost penalties for each model-task pair; and (3) final assignments with their objective scores.
For each routing decision, an LLM transforms the log into concise explanations by identifying which skills drove model selection for high-complexity tasks, explaining cost-quality tradeoffs when cheaper models were selected despite lower capabilities, and linking decisions to user preferences. Because explanations are derived from real match scores and cost penalties, they reflect actual decision logic rather than post-hoc rationalization. This approach enables developers to verify that routing decisions correctly balance user preferences against model capabilities and costs. We provide all prompts for \topaz~in Appendix~\ref{apdx:prompts}.

We note that routing is one layer of a multi-faceted agentic stack: \topaz~explains \textit{which model was selected and why}, not the downstream behavior of the selected model's output. Monitoring actual model inputs and outputs to assess the downstream effects of routing on end-to-end agent performance is an important direction for future work.
%
%
%

\textbf{Local and Global Explanations.} \topaz's explanation framework provides both local and global explanations, as different stakeholders benefit from different granularities: a developer debugging a single failure needs local explanations, while a product manager evaluating overall cost-quality tradeoffs needs global ones. Per-task explanations serve as \textit{local} rationale: why a specific model was selected for a specific subtask given its skill requirements and cost constraints. Cross-task summaries---such as those in Table~\ref{tab:routing_results}---act as \textit{global} explanations that characterize the router's overall strategy through aggregating local justifications into higher-level strategies. This distinction becomes especially important as workflows scale to dozens of subtasks and local explanations become impractical to review individually.

\textbf{Feedback Loop.} To incorporate developer preferences into routing decisions, we introduce a closed-loop feedback mechanism that allows users to adjust $q_t$, originally LLM-profiled. If quality was insufficient for a specific sub-task, increasing $q_t$ for similar future tasks shifts the cost-quality balance for future routing decisions. \label{sec:feedback_loop}

\section{System Demonstration and Case Study}



\subsection{Experimental Setup}

\begin{table}[b]
    \small
    \centering
    \caption{Model skill profiles from \topaz~ and costs per million tokens (USD).}
    \label{tab:skill_profles}
    \renewcommand{\arraystretch}{1.2} 
    \begin{tabular}{lcccccccccc}
        \toprule
        \textbf{Model} & \textbf{Math} & \textbf{Logic} & \textbf{Code} & \textbf{Tool} & \textbf{Fact.} & \textbf{Write} & \textbf{Instr.} & \textbf{Summ.} & \textbf{$p^{in}$(\$)} & \textbf{$p^{out}$(\$)} \\
        \midrule
        Claude-Opus-4.5 & .967 & .966 & .974 & \cellcolor{green!20}.988 & .955 & .969 & .979 & .963 & \cellcolor{red!15}5.00 & \cellcolor{red!15}25.00 \\
        Gemini-3-Pro & \cellcolor{green!20}.999 & \cellcolor{green!20}.988 & .981 & .953 & \cellcolor{green!20}.999 & \cellcolor{green!20}.999 & \cellcolor{green!20}.984 & \cellcolor{green!20}.996 & 2.00 & 12.00 \\
        GPT-5.2 & .991 & .974 & \cellcolor{green!20}.992 & .849 & .981 & .971 & .903 & .995 & 1.75 & 14.00 \\
        Llama-4-Maverick & .660 & .626 & \cellcolor{red!20}.433 & \cellcolor{red!20}.504 & .826 & .851 & \cellcolor{red!20}.719 & \cellcolor{red!20}.817 & 0.15 & 0.60 \\
        Mistral-Small-3.1 & \cellcolor{red!20}.506 & \cellcolor{red!20}.578 & .593 & .544 & \cellcolor{red!20}.704 & .872 & .763 & \cellcolor{red!20}.817 & \cellcolor{green!20}0.10 & \cellcolor{green!20}0.30 \\
        \bottomrule
    \end{tabular}
\end{table}

\textbf{Models.} We compare five models spanning the cost-capability spectrum: Gemini 3 Pro, Claude Opus 4.5, GPT 5.2, Llama 4 Maverick, and Mistral Small 3.1~\citep{gemini3pro2025, claudeopus45_2025, gpt52openai2025, llama4maverickhf2025, mistralsmall31_2025}.\footnote{Token prices retrieved from Anthropic, Google, OpenAI, and OpenRouter.} We use Gemini 3.0 Flash \cite{gemini3flash2025} to profile benchmarks and tasks and to generate explanations.
\quad \textbf{Benchmarks.} We assess models across diverse capabilities using: TextArena~\citep{chiang2024lmarena} and Search Arena~\citep{miroyan2025searcharena} for conversational quality and retrieval; BFCL v4~\citep{yan2024bfcl} for tool-use; SWE-bench~\citep{jimenez2023swebench} and LiveCodeBench~\citep{jain2024livecodebench} for software engineering; MMMU~\citep{yue2023mmmu} for multimodal reasoning; GPQA~\citep{rein2023gpqa} and MMLU-Pro~\citep{wang2024mmlupro} for domain knowledge; and MATH-500~\citep{math500hf} and AIME~\citep{aime2024} for mathematical reasoning. Benchmark scores are pulled from public leaderboard sites in February 2026 (see Appendix~\ref{apdx:benchmark_scores}).
\quad \textbf{Skills.} Each model was profiled across eight skills: mathematical reasoning, logical reasoning, code generation, tool use, factual knowledge, writing quality, instruction following, and summarization. Further details are in Appendix~\ref{apdx:models_benchmarks_skills}. We profiled model capabilities across these skills following Eq. \ref{eq:capability}, with results in Table \ref{tab:skill_profles} revealing a spectrum of abilities. Since downstream explanations are only as meaningful as the underlying skill taxonomy, we place additional emphasis on precise and auditable profiling for models.

\begin{figure}[t]
    \centering
    \includegraphics[width=1\linewidth]{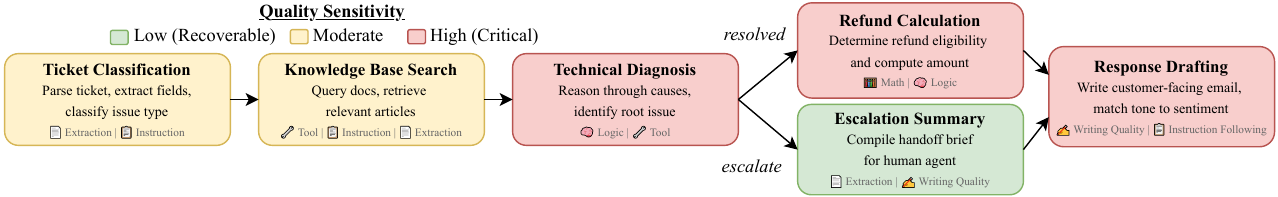}
    \Description{Flowchart of a six-stage customer support pipeline. Ticket Classification (low sensitivity, green) feeds into Knowledge Base Search (moderate sensitivity, yellow), which feeds into Technical Diagnosis (high sensitivity, red). Technical Diagnosis branches into two paths: resolved leads to Refund Calculation (high sensitivity, red) then Response Drafting (high sensitivity, red); escalate leads to Escalation Summary (low sensitivity, green). Each stage lists its required skills, such as logic and tool use for Technical Diagnosis, and math and logic for Refund Calculation.}
    \caption{\textbf{Customer Support Agentic Workflow} with colors representing quality sensitivity. The AI Agent parses and classifies tickets, queries relevant documents, and reasons to form a technical diagnosis, lastly escalating or directly responding to the customer.}
    \vspace{-2 ex}
    \label{fig:support_pipeline}
\end{figure}

\vspace{-2 ex}
\subsection{Case Study: Customer Support Escalation}

We demonstrate \topaz~on a customer support pipeline that processes tickets from intake through resolution or human escalation (Figure~\ref{fig:support_pipeline}). Our results showcase \topaz's explainable routing with varied task complexity and skill demands. Due to limited space, we focus our case study only on objective-based routing and explanations. See Appendix  \ref{apdx:budget_routing} for an example of budget-based routing.

\textbf{Pipeline Configuration.} The pipeline consists of six tasks with developer-configurable quality sensitivities. For instance, \textit{Technical Diagnosis} ($q=1.0$) demands high accuracy to avoid wasting engineering time on incorrect diagnoses, while \textit{Escalation Summary} ($q=0.4$) tolerates lower quality as it precedes internal human review.

\newcommand{\assign}[2]{\makebox[0.4cm][l]{#1}→#2}
\begin{table*}[h]
\setlength{\fboxsep}{2pt}
\vspace{-8pt}
\small
\centering
\caption{Routing assignments and system-generated explanations across cost sensitivity settings.}
\vspace{-10pt}
\label{tab:routing_results}
\setlength{\tabcolsep}{4pt}
\begin{tabular}{c|l|p{11cm}}
\toprule
\textbf{$c_{\text{global}}$} & \textbf{Assignments} & \textbf{Trace-driven Explanation of Routing Decisions} \\
\midrule
\textbf{0.00} & 
\begin{tabular}[t]{@{}l@{}}
\assign{TC}{ Gemini-3-Pro} \\
\assign{KB}{ Claude-Opus-4.5} \\
\assign{TD}{ Claude-Opus-4.5} \\
\assign{RC}{ Gemini-3-Pro} \\
\assign{RD}{ Gemini-3-Pro} \\
\assign{ES}{ Gemini-3-Pro}
\end{tabular}
& With cost sensitivity set to zero, the engine prioritized raw performance over cost, selecting Claude Opus for the most complex tasks—Technical Diagnosis and Knowledge Base Search—due to its superior logical reasoning and tool-use capabilities. Gemini 3 Pro was routed to tasks like Refund Calculation and Response Drafting where its high-tier mathematical reasoning and writing quality were the primary success factors. For simpler objectives like Ticket Classification and Escalation Summary, the system selected Gemini because it fully satisfied the required skill sets [$\ldots$]\\

\midrule
\textbf{0.50} & 
\begin{tabular}[t]{@{}l@{}}
\assign{TC}{\colorbox{green!15}{Mistral-Small-3.1}} \\
\assign{KB}{\colorbox{green!15}{Gemini-3-Pro}} \\
\assign{TD}{\colorbox{green!15}{Gemini-3-Pro}} \\
\assign{RC}{ Gemini-3-Pro} \\
\assign{RD}{ Gemini-3-Pro} \\
\assign{ES}{\colorbox{green!15}{Mistral-Small-3.1}}
\end{tabular}
& For high-complexity tasks like Technical Diagnosis and Refund Calculation, Gemini was selected because it offered the strongest alignment with critical skills like logical reasoning and mathematical precision while maintaining a significantly better cost-to-performance ratio than Claude. In contrast, for simpler or high-volume tasks such as Ticket Classification and Escalation Summary, the engine prioritized Mistral to maximize cost savings, as its summarization and extraction capabilities were sufficient for the lower quality-sensitivity requirements. Ultimately, the system reserves premium compute for tasks requiring specialized skills like tool use [$\ldots$] \\

\midrule
\textbf{1.00} & 
\begin{tabular}[t]{@{}l@{}}
\assign{TC}{\colorbox{green!15}{Mistral-Small-3.1}} \\
\assign{KB}{\colorbox{orange!15}{Mistral-Small-3.1}} \\
\assign{TD}{ Gemini-3-Pro} \\
\assign{RC}{\colorbox{orange!15}{Mistral-Small-3.1}} \\
\assign{RD}{\colorbox{orange!15}{Mistral-Small-3.1}} \\
\assign{ES}{\colorbox{green!15}{Mistral-Small-3.1}}
\end{tabular}
& For most tasks, including Ticket Classification and Response Drafting, the engine prioritized Mistral because your high cost sensitivity outweighed the marginal quality gains of larger models, as Mistral provided sufficient capabilities for summarization and writing. However, for the high-complexity Technical Diagnosis task, the system selected Gemini  to ensure the superior logical reasoning and tool-use performance necessary for that specific objective. Even for difficult tasks like Refund Calculation, the engine opted for more economical models like Mistral, effectively trading off specialized mathematical reasoning to adhere to your strict cost constraints.\\

\bottomrule

\multicolumn{3}{l}{\begin{minipage}{\textwidth}
\vspace{0.5mm}
\footnotesize TC=Ticket Classification, KB=Knowledge Base Search, TD=Technical Diagnosis, RC=Refund Calculation, RD=Response Drafting, ES=Escalation Summary. See Appendix~\ref{apdx:full_results_from_routing} for full score breakdowns and results at $c_{\text{global}} = 0.05,0.95$.
\end{minipage}}
\end{tabular}
\vspace{-4mm}
\end{table*}

\vspace{1 ex}
\textbf{Routing Behavior Across Cost Sensitivities.} Table~\ref{tab:routing_results} demonstrates how \topaz~adapts routing under three cost configurations while providing trace-driven explanations. At $c_{\text{global}} = 0.0$ (performance-optimal), \topaz~assigns Claude to complex diagnosis and tool-heavy search because it best matches the required reasoning and tool-use skills, and routes remaining tasks to Gemini for its strength in math and writing. At $c_{\text{global}} = 0.5$ (balanced), \topaz~replaces Claude with Gemini for Technical Diagnosis---explaining that Gemini offers comparable skill coverage at less than half the cost---and downgrades extraction tasks to Mistral, whose capabilities fully satisfy those tasks' lower skill requirements due to their lesser complexity. At $c_{\text{global}} = 1.0$ (cost-optimal), \topaz~retains Gemini only for Technical Diagnosis, where the task's high complexity demands strong logical reasoning that cheaper models cannot meet, and assigns Mistral everywhere else. Across all three settings, the generated explanations let developers verify that \topaz's cost savings stem from capability saturation rather than hidden quality loss, and pinpoint which tasks are most sensitive to further budget changes due to their importance to workflow success.


\section{Conclusions}

We present \topaz, an inherently interpretable model router for agentic workflows that grounds every assignment in human-interpretable skill profiles, traceable cost-quality optimization, and natural-language explanations derived from actual routing traces. 
Our case study demonstrates that \topaz~adapts coherently across budgetary preferences while enabling developers to audit, diagnose, and steer the routing process with fine-grained control. 
As AI increasingly shifts away from monolithic models toward complex, multi-agent architectures, balancing economic realities with operational transparency will be critical for real-world deployment. By bridging the gap between cost-aware routing and actionable explainability, this approach establishes a necessary foundation for trustworthy AI orchestration.
We hope our work motivates further research into human-centered transparency for scalable, multi-model agentic systems.


\newpage
\bibliographystyle{ACM-Reference-Format}
\bibliography{sample-base}

\appendix
\newpage
\section{Prompts for \topaz}
\label{apdx:prompts}

All prompts use Gemini 3.0 Flash as the profiling LLM. Template variables are shown in \texttt{\{braces\}}. Skill taxonomy definitions (omitted for brevity) enumerate the eight skills from Section~\ref{section:skill_profiles} with natural-language descriptions.

\subsection{Prompt: Benchmark to Skills}
\label{apdx:prompt_benchmark}

This prompt profiles a benchmark to determine which skills it primarily measures. The output is an $L_1$-normalized skill weight vector used to compute model capability profiles.

\begin{promptbox}[Benchmark Skill Profiling Prompt]
\small
You are profiling an LLM benchmark. Determine what skills this benchmark primarily measures. A benchmark may test multiple skills, but focus on the dominant capabilities it evaluates. What skills are the most relevant for the LLM to possess in order to perform well on this benchmark?

\medskip
\textbf{SKILL TAXONOMY} (assign weights from these and ONLY these):\\
\texttt{\{skill\_definitions\}}

\medskip
\textbf{Benchmark:} \texttt{\{benchmark\_name\}}\\
\textbf{Description:} \texttt{\{benchmark\_description\}}\\
\texttt{\{example\_items\_block\}} \hfill \textit{(optional, up to 5 samples)}

\medskip
\textbf{INSTRUCTIONS:}
\begin{enumerate}[leftmargin=*, nosep]
    \item Assign weights summing to 1.0 across all skills according to their importance for this benchmark.
    \item Set skills to 0.0 if they are NOT required or measured. Be aggressive about zeroing out irrelevant skills --- most benchmarks involve only 2--4 skills.
    \item For each nonzero skill, provide a one-sentence rationale.
\end{enumerate}

\medskip
Respond with ONLY a JSON object in this exact format:\\
\texttt{\{"skill\_weights": \{"<skill>": <float>, ...\},}\\
\texttt{\ "rationale": \{"<skill>": "<justification>", ...\}\}}
\end{promptbox}

\subsection{Prompt: Subtask to Skills}
\label{apdx:prompt_task}

Each subtask in an agentic workflow is profiled independently for skill requirements, then all subtasks are jointly analyzed for relative complexity and quality sensitivity.

\begin{promptbox}[Subtask Skill Profiling Prompt]
\small
You are profiling a subtask in an agentic AI pipeline. Determine what skills an LLM needs to perform this task well. This task may tangentially require many skills, but focus on the dominant skills required to complete this task successfully. Focus on the capabilities that most strongly determine success or failure.

\medskip
\textbf{SKILL TAXONOMY} (assign weights from these and ONLY these):\\
\texttt{\{skill\_definitions\}}

\medskip
\textbf{Task:} \texttt{\{task\_name\}}\\
\textbf{Description:} \texttt{\{task\_description\}}

\medskip
\textbf{INSTRUCTIONS:}
\begin{enumerate}[leftmargin=*, nosep]
    \item Assign weights summing to 1.0 across all skills according to their importance for this task.
    \item Set skills to 0.0 if they are NOT required. Be aggressive about zeroing out irrelevant skills --- most tasks involve only 2--4 skills.
    \item For each nonzero skill, provide a one-sentence rationale.
\end{enumerate}

\medskip
Respond with ONLY a JSON object in this exact format:\\
\texttt{\{"skill\_weights": \{"<skill>": <float>, ...\},}\\
\texttt{\ "rationale": \{"<skill>": "<justification>", ...\}\}}
\end{promptbox}

\medskip

After individual profiling, all subtasks are jointly analyzed to extract relative metadata:

\begin{promptbox}[Pipeline-Relative Metadata Prompt]
\small
You are profiling an agentic AI pipeline. Below are ALL subtasks in this pipeline. Your job is to assess each subtask's metadata RELATIVE TO THE OTHER TASKS in this pipeline. This is critical: scores should reflect how tasks compare to each other, not in isolation.

\medskip
\texttt{\{subtask\_list\}} \hfill \textit{(name and description for each subtask)}

\medskip
For EACH subtask, provide:
\begin{enumerate}[leftmargin=*, nosep]
    \item \textbf{complexity} (float, 0--1): How difficult is this task? Score RELATIVE to the other tasks: the hardest task should be near the top, the easiest near the bottom.
    \item \textbf{quality\_sensitivity} (float, 0--1): How important is it to get this task right vs.\ saving cost? 1.0 = errors are costly, hard to detect, or propagate downstream.
    \item \textbf{estimated\_input\_tokens} (int): Approximate input tokens including context from prior stages.
    \item \textbf{estimated\_output\_tokens} (int): Approximate output tokens for the task's deliverable.
    \item \textbf{rationale} (string): One sentence explaining the relative positioning.
\end{enumerate}

\medskip
Be aggressive about differentiation. A data cleaning step is NOT as complex as multi-step analytical reasoning.

\medskip
Respond with ONLY a JSON object mapping each task name to its metadata.
\end{promptbox}

\subsection{Prompt: Developer Explanation}
\label{apdx:prompt_explanation}

After routing, \topaz{} synthesizes the numerical routing trace into a natural-language explanation. The explanation prompt receives the full structured explanation log, which contains: user configuration (cost sensitivity $c_\text{global}$, per-task quality sensitivities $q_t$), per-model scoring details for each task (skill match scores, cost penalties, final objective scores), and the winning model assignment for each subtask.

\begin{promptbox}[Routing Explanation Prompt]
\small
You are an expert AI systems analyst. Given the following explainability log from a model selection engine, generate a human-readable and short explanation of how models were selected for tasks based on user preferences, model capabilities, and cost considerations.

\medskip
Your audience is a developer of the system. They want to understand why this model was chosen for this specific task over the other models. Tie this explanation back into the tradeoffs between quality and cost, and how the user's preferences influenced the decision.

\medskip
Do not mention the formulas, variables, or raw numbers. Instead, focus on the high-level reasoning process and the key factors that led to the selection. Speak to the intuition behind the decisions, rather than the technical details. Do not be overly verbose or simply summarize the log. Instead, synthesize a high-level explanation of the overall decision process.

\medskip
For example, discuss how a certain, more complex task might require a model with higher capabilities, so a model strong at a specific skill required for that task was selected even if it was more expensive, because the user indicated that quality was more important for that task. Or how a simpler task with less stringent quality requirements ended up being assigned to a cheaper model. Mention the specific models, tasks, and skills where relevant. If you chose a model because high performance was needed, cite what skill(s) are most relevant and how the selected model excels at them. If you had to cut costs and select a model with lower capabilities, explain this tradeoff.

\medskip
Keep your explanation concise (3--4 sentences). Speak as if you are talking to the creator of this agent.

\medskip
\textbf{Data:} \texttt{\{explain\_log\}}
\end{promptbox}

\section{Elaboration on Case Study}

In our case study, we utilized the \topaz~system to analyze an customer support agent and provide model recommendations under several models. Our full experimental setup and results are below.

\subsection{Models, Benchmarks, and Skills}
\label{apdx:models_benchmarks_skills}

For our experimental evaluation, we synthesize model capability profiles across a diverse set of models, explained in detail in Table \ref{tab:models}, and benchmarks, elaborated on in Table \ref{tab:benchmarks}. We analyze the benchmarks across 8 skills: mathematical reasoning, logical reasoning, code generation, tool use, factual knowledge, writing quality, instruction following, and summarization. 

\paragraph{Capability Score Calibration.}
\label{apdx:calibration}
Standardized benchmarks evaluate upper-bound model performance through high-complexity stress testing, while practical agentic subtasks typically impose median-utility requirements. We apply a calibration factor $\kappa = 0.2$ to the raw capability scores, yielding $C'_{m,s} = \kappa \cdot C_{m,s}$, mapping the high-ceiling benchmark space onto the operational requirement space of our task set so that model capabilities are evaluated relative to task needs rather than absolute theoretical limits. Without this calibration, the satisfaction ratio $C_{m,s} / (k_t \cdot R_{t,s})$ saturates at 1.0 for most model-task pairs, collapsing the routing signal and preventing meaningful differentiation between models. The value $\kappa = 0.2$ was selected through manual tuning to maximize discriminability across model-task assignments; specifically, we swept $\kappa \in \{0.1, 0.2, 0.3, 0.5, 0.6, 0.7, 0.8, 0.9, 1.0\}$ and selected the smallest value at which skill match scores differentiated meaningfully across all model-task pairs without saturating at 1.0 for more than one model per task. We found this approach more auditable and transparent than learned normalization.

\begin{table}[ht]
\small
\centering
\caption{Language Model Specifications and Pricing}
\label{tab:models}
\begin{adjustbox}{max width=\textwidth}
\begin{tabular}{l l r r p{7cm}}
\toprule
\textbf{Model} & \textbf{Provider} & \textbf{Cost In} & \textbf{Cost Out} & \textbf{Description} \\
& & \textbf{(\$/M)} & \textbf{(\$/M)} & \\
\midrule
Claude Opus 4.5 & Anthropic & 5.00 & 25.00 & Most intelligent Claude model combining maximum capability with practical performance, with improvements in reasoning, coding, and complex problem-solving for agentic workflows \\
\midrule
Gemini 3 Pro & Google & 2.00 & 12.00 & Multimodal reasoning model with 1M context window and dynamic thinking levels, best for complex tasks requiring broad world knowledge and advanced reasoning \\
\midrule
GPT-5.2 & OpenAI & 1.75 & 14.00 & Flagship model for professional knowledge work with adaptive reasoning, with improvements in long-context understanding, agentic tool calling, and artifact creation\\
\midrule
Llama 4 Maverick & Meta & 0.15 & 0.60 & Open-weight natively multimodal MoE model with 1M context, designed for advanced reasoning, multilingual chat, image understanding, and code generation at high cost efficiency \\
\midrule
Mistral Small 3.1 & Mistral AI & 0.10 & 0.30 & Open-weight multimodal model designed for fast conversational assistance, low-latency function calling, and fine-tuning into domain-specific experts on consumer hardware \\
\bottomrule
\end{tabular}
\end{adjustbox}
\end{table}

\begin{table}[ht]
\small
\centering
\caption{Benchmark Specifications}
\label{tab:benchmarks}
\begin{adjustbox}{max width=\textwidth}
\begin{tabular}{l p{12cm}}
\toprule
\textbf{Benchmark} & \textbf{Description} \\
\midrule
TextArena & Open platform for evaluating LLMs via pairwise human preference voting with 240K+ votes and Elo ratings. Crowdsourced prompts span diverse open-ended tasks such as drafting letters, creative writing, and general assistance. \\
\midrule
SearchArena & 24K multi-turn interactions with 12K human preference votes evaluating search-augmented LLMs. Tests whether models can effectively integrate web search results and citations into responses for queries requiring current or niche factual information. \\
\midrule
BFCL v4 & Evaluates function calling ability using AST-based evaluation across serial and parallel invocations in multiple programming languages. Includes stateful multi-step agentic settings that test memory, dynamic decision-making, and abstention. \\
\midrule
SWE-bench & 2,294 real software engineering problems drawn from GitHub issues and pull requests across 12 popular Python repositories. Models must navigate large codebases and generate multi-file patches that resolve actual bugs and feature requests. \\
\midrule
LiveCodeBench & Contamination-free coding benchmark that continuously collects competitive programming problems from LeetCode, AtCoder, and CodeForces. Evaluates code generation, self-repair, and execution prediction on algorithmic challenges. \\
\midrule
MMMU & 11.5K college-level multimodal questions from exams and textbooks spanning 30 subjects and 183 subfields. Requires interpreting heterogeneous image types including charts, diagrams, chemical structures, and music sheets alongside domain-specific reasoning. \\
\midrule
GPQA & 448 expert-written multiple-choice questions in biology, physics, and chemistry at graduate level. Designed to be ``Google-proof'': PhD experts reach 65\% accuracy while skilled non-experts achieve only 34\% even with unrestricted web access. \\
\midrule
MMLU-Pro & Reasoning-focused extension of MMLU with 10-choice questions that are 16--33\% harder than the original. Eliminates trivial questions and rewards chain-of-thought reasoning across 14 diverse subject areas. \\
\midrule
MATH-500 & 500 held-out competition mathematics problems sampled from the 12.5K MATH dataset. Covers challenging high-school competition topics with full step-by-step solutions requiring rigorous derivations. \\
\midrule
AIME 2024 & Prestigious invite-only competition for top 5\% AMC scorers, with 15 problems of increasing difficulty across algebra, geometry, number theory, and combinatorics. Each answer is a single integer from 0 to 999. \\
\bottomrule
\end{tabular}
\end{adjustbox}
\end{table}

\subsection{Benchmark Scores}
\label{apdx:benchmark_scores}

Model scores for each benchmark were collected from publicly available leaderboards and evaluation platforms. Raw scores used to compute the capability profiles in Table~\ref{tab:skill_profles} are sourced from the following:

\begin{itemize}
    \item \textbf{Vals.ai} (\url{https://www.vals.ai/benchmarks}): AIME 2024, MATH-500, GPQA, MMLU-Pro, MMMU, SWE-Bench, LiveCodeBench
    \item \textbf{Berkeley Function Calling Leaderboard} (\url{https://gorilla.cs.berkeley.edu/leaderboard.html}): BFCL v4
    \item \textbf{Arena.ai} (\url{https://arena.ai/}): TextArena and SearchArena Elo ratings, as of February 2026
    \item \textbf{LLM Stats} (\url{https://llm-stats.com/benchmarks}): Cross-referenced metrics for some models for greater model coverage
\end{itemize}

Where a model appeared on multiple leaderboards, we used the most recent reported score. All scores were normalized per benchmark using 0-max normalization as described in Section~\ref{section:skill_profiles}. 

\subsection{Benchmark Skill Profiles}
\label{apdx:benchmark_profile_skill}

Table~\ref{tab:benchmark_profiles} shows the skill weight decompositions assigned to each benchmark. These weights are determined by prompting the LLM profiler with each benchmark's description and example items (Appendix~\ref{apdx:prompt_benchmark}), yielding an $L_1$-normalized distribution over the eight skills. Benchmarks are generally sparse: most assign nonzero weight to a few skills. These weights are used directly in Eq.~\ref{eq:capability} to compute model capability profiles.

\begin{table*}[ht]
\centering
\small
\caption{Benchmark skill weight decompositions used by \topaz{} to compute model capability profiles. Each row defines how a benchmark's score is attributed across skills, based on analysis of the benchmark's design, question types, and evaluation criteria.}
\label{tab:benchmark_profiles}
\setlength{\tabcolsep}{3.5pt}
\begin{tabular}{l cccccccc c}
\toprule
& \multicolumn{8}{c}{\textbf{Skill Weights ($w$)}} & \\
\cmidrule(lr){2-9}
\textbf{Benchmark} & \textbf{Math} & \textbf{Logic} & \textbf{Code} & \textbf{Tool} & \textbf{Fact.} & \textbf{Write} & \textbf{Instr.} & \textbf{Summ.} & \textbf{Max Score} \\
\midrule
TextArena & --- & .15 & --- & --- & .15 & .35 & .35 & --- & 1481 \\
SearchArena & --- & --- & --- & .30 & .20 & .20 & --- & .30 & 1224 \\
BFCL v4 & --- & .15 & --- & .70 & --- & --- & .15 & --- & 77.47 \\
SWE-bench Verified & --- & .40 & .30 & .30 & --- & --- & --- & --- & 75.4 \\
LiveCodeBench & .10 & .30 & .50 & --- & --- & --- & .10 & --- & 86.41 \\
MMMU & .30 & .30 & --- & --- & .40 & --- & --- & --- & 87.63 \\
GPQA Diamond & .20 & .45 & --- & --- & .35 & --- & --- & --- & 91.67 \\
MMLU-Pro & .30 & .30 & --- & --- & .40 & --- & --- & --- & 90.1 \\
MATH-500 & .70 & .30 & --- & --- & --- & --- & --- & --- & 96.4 \\
AIME 2024 & .70 & .30 & --- & --- & --- & --- & --- & --- & 96.88 \\
\bottomrule
\multicolumn{10}{l}{\footnotesize Dashes indicate zero weight. Max Score is the highest score achieved by any model on the benchmark.}
\end{tabular}
\end{table*}
\subsection{Customer Support Subtask Profiles}
\label{apdx:capability_profile_model_skill}
Table~\ref{tab:subtask_profiles} details the skill requirement profiles, complexity scores, quality sensitivity values, and token estimates for each subtask in the customer support pipeline case study. These profiles are generated by the subtask profiling prompts in Appendix~\ref{apdx:prompt_task} and serve as the task-side input to the routing objective.

\begin{table*}[ht]
\centering
\small
\caption{Customer support pipeline subtask profiles. Skill requirements ($R_{t,s}$) define the required capability distribution; complexity ($k$) and quality sensitivity ($q$) parameterize the routing objective. Token count estimates $\sigma_{\text{in}}, \sigma_{\text{out}}$ determine per-call cost.}
\label{tab:subtask_profiles}
\setlength{\tabcolsep}{3.5pt}
\begin{tabular}{l cccccccc cc cc}
\toprule
& \multicolumn{8}{c}{\textbf{Skill Requirements ($R_{t,s}$)}} & \multicolumn{2}{c}{\textbf{Routing Params}} & \multicolumn{2}{c}{\textbf{Token Est.}} \\
\cmidrule(lr){2-9} \cmidrule(lr){10-11} \cmidrule(lr){12-13}
\textbf{Subtask} & \textbf{Math} & \textbf{Logic} & \textbf{Code} & \textbf{Tool} & \textbf{Fact.} & \textbf{Write} & \textbf{Instr.} & \textbf{Summ.} & \textbf{$q$} & \textbf{$k$} & \textbf{$\sigma_{\text{in}}$} & \textbf{$\sigma_{\text{out}}$} \\
\midrule
Ticket Classification & ---  & .10  & ---  & ---  & ---  & ---  & .40  & .50  & 0.65 & 0.25 &  400 &   80 \\
Knowledge Base Search & ---  & ---  & ---  & .40  & ---  & ---  & .30  & .30  & 0.55 & 0.50 &  500 & 1000 \\
Technical Diagnosis   & ---  & .40  & ---  & .30  & ---  & ---  & .10  & .20  & 1.00 & 0.95 & 2000 &  500 \\
Refund Calculation    & .40  & .40  & ---  & ---  & ---  & ---  & .20  & ---  & 0.95 & 0.80 & 1200 &  200 \\
Response Drafting     & ---  & ---  & ---  & ---  & ---  & .60  & .40  & ---  & 0.90 & 0.60 & 1500 &  400 \\
Escalation Summary    & ---  & .10  & ---  & ---  & ---  & .20  & .20  & .50  & 0.40 & 0.30 & 3000 &  250 \\
\bottomrule
\multicolumn{13}{l}{\footnotesize Dashes indicate zero weight.}
\end{tabular}
\end{table*}

\subsection{Complete Results from Dual-Objective Routing}
\label{apdx:full_results_from_routing}

Table~\ref{tab:full_routing_app} reports complete routing decisions across all five cost sensitivity settings ($c_{\text{global}} \in \{0.00, 0.05, 0.50, 0.95, 1.00\}$) for each subtask in the customer support pipeline. Each entry shows the selected model and its runner-up, along with their respective skill match scores ($M$), normalized cost penalties ($C$), and final objective scores ($S$), with the margin $\Delta$ over the runner-up indicating decision confidence. At $c_{\text{global}} = 0.00$, cost is ignored entirely and quality-maximizing models are selected; Claude Opus and Gemini-3 Pro dominate for tasks requiring strong logical reasoning or tool use. As cost sensitivity increases, cheaper models (Mistral Small 3.1, Llama 4 Maverick) win tasks where their match scores are competitive or where quality sensitivity is low. Notably, Gemini-3 Pro persists across most settings for Technical Diagnosis and Refund Calculation due to its strong match score edge, noting how skill alignment is preserved for critical tasks even at high cost sensitivity.

\begin{table*}[t]
\centering
\small
\caption{Routing decisions across cost sensitivity settings showing the selected model, runner-up, and the dominant factor in each decision. $M$ = skill match (dot product), $C$ = normalized cost penalty, $S$ = final objective score, $\Delta$ = score margin over runner-up.}
\label{tab:full_routing_app}
\setlength{\tabcolsep}{4pt}
\begin{tabular}{c l lrrr lrrr r p{3.4cm}}
\toprule
& & \multicolumn{4}{c}{\textbf{Selected Model}} & \multicolumn{4}{c}{\textbf{Runner-up}} & & \\
\cmidrule(lr){3-6} \cmidrule(lr){7-10}
$c_{\text{global}}$ & \textbf{Task} & \textbf{Model} & $M$ & $C$ & $S$ & \textbf{Model} & $M$ & $C$ & $S$ & $\Delta$ & \textbf{Decisive Factor} \\
\midrule
\multicolumn{12}{l}{\cellcolor{gray!10}\textbf{$c_{\text{global}} = 0.00$}: quality-only (cost weight $0$ for all tasks)} \\
\midrule
  \multirow{6}{*}{\textbf{0.00}} & TC & Gemini-3 & 1.00 & .143 & .650 & \emph{all tied} & 1.00 & 0.143 & .650 & .000 & Tiebreak best non-capped $M$ \\
   & KB & Claude-O & .995 & 1.00 & .547 & Gemini-3 & .981 & .154 & .540 & .007 & Highest $M$ via tool\_use \\
   & TD & Claude-O & .711 & 1.00 & .711 & Gemini-3 & .709 & .144 & .709 & .002 & Highest $M$ in logic+tool\_use \\
   & RC & Gemini-3 & .697 & .142 & .662 & GPT-5.2 & .691 & .145 & .657 & .005 & Best math+logic $M$ \\
   & RD & Gemini-3 & .661 & .145 & .595 & Claude-O & .649 & 1.00 & .584 & .011 & Best writing+instr.\ $M$ \\
   & ES & Gemini-3 & 1.00 & .137 & .400 & \emph{all tied} & 1.00 & .00 & .400 & .000 & Tiebreak best non-capped $M$  \\
\midrule
\multicolumn{12}{l}{\cellcolor{gray!10}\textbf{$c_{\text{global}} = 0.05$}: near quality-only (minimal cost influence)} \\
\midrule
  \multirow{6}{*}{\textbf{0.05}} & TC & Mistral & 1.00 & .000 & .618 & Llama-4 & 1.00 & .009 & .617 & .001 & $M$ tied at 1; lowest $C$ wins \\
   & KB & Gemini-3 & .981 & .154 & .509 & Claude-O & .995 & 1.00 & .498 & .011 & $M$ outweighs small $C$ weight \\
   & TD & Claude-O & .711 & 1.00 & .675 & Gemini-3 & .709 & .144 & .673 & .002 & $M$ outweighs small $C$ weight \\
   & RC & Gemini-3 & .697 & .142 & .629 & GPT-5.2 & .691 & .145 & .624 & .005 & Best $M$; cost negligible \\
   & RD & Gemini-3 & .661 & .145 & .564 & Claude-O & .649 & 1.00 & .550 & .014 & Best $M$ \& better $C$\\
   & ES & Mistral & 1.00 & .000 & .380 & Llama-4 & 1.00 & .009 & .380 & .000 & $M$ tied at 1; lowest $C$ wins \\
\midrule
\multicolumn{12}{l}{\cellcolor{gray!10}\textbf{$c_{\text{global}} = 0.50$}: balanced (cost differentiates when $q < 1$)} \\
\midrule
  \multirow{6}{*}{\textbf{0.50}} & TC & Mistral & 1.00 & .000 & .325 & Llama-4 & 1.00 & .009 & .324 & .001 & $M$ tied; lowest $C$ wins \\
   & KB & Gemini-3 & .981 & .154 & .235 & Mistral & .818 & .000 & .225 & .010 & $M$ edge outweighs $C$ gap \\
   & TD & Gemini-3 & .709 & .144 & .354 & Claude-O & .711 & 1.00 & .351 & .003 & $C$ penalty outweighs $M$ score \\
   & RC & Gemini-3 & .697 & .142 & .328 & GPT-5.2 & .691 & .145 & .325 & .003 & Better $M$ for cheaper $C$  \\
   & RD & Gemini-3 & .661 & .145 & .290 & GPT-5.2 & .625 & .153 & .273 & .017 & $M$ gap too large for $C$ \\
   & ES & Mistral & 1.00 & .000 & .200 & Llama-4 & 1.00 & .009 & .197 & .003 & $M$ tied; lowest $C$ wins \\
\midrule
\multicolumn{12}{l}{\cellcolor{gray!10}\textbf{$c_{\text{global}} = 0.95$}: cost-dominant (quality weight small but above $\varepsilon$ floor)} \\
\midrule
  \multirow{6}{*}{\textbf{0.95}} & TC & Mistral & 1.00 & .000 & .033 & Llama-4 & 1.00 & .009 & .030 & .003 & Cost dominates; cheapest wins \\
   & KB & Mistral & .818 & .000 & .022 & Llama-4 & .789 & .008 & .018 & .004 & Cost dominates; cheapest wins \\
   & TD & Gemini-3 & .709 & .144 & .034 & GPT-5.2 & .684 & .152 & .033 & .001 & $M$ quality weight decisive\\
   & RC & Gemini-3 & .697 & .142 & .026 & GPT-5.2 & .691 & .145 & .026 & .000 & $M$ quality weight decisive \\
   & RD & Mistral & .545 & .000 & .025 & Llama-4 & .523 & .009 & .023 & .002 & Cost dominates, cheapest wins \\
   & ES & Mistral & 1.00 & .000 & .020 & Llama-4 & 1.00 & .009 & .015 & .005 & Cost dominates; cheapest wins \\
\midrule
\multicolumn{12}{l}{\cellcolor{gray!10}\textbf{$c_{\text{global}} = 1.00$}: cost-minimizing (quality weight $= q_t \cdot \varepsilon$)} \\
\midrule
  \multirow{6}{*}{\textbf{1.00}} & TC & Mistral & 1.00 & .000 & .007 & Llama-4 & 1.00 & .009 & .003 & .004 & Cost dominates, cheapest wins \\
   & KB & Mistral & .818 & .000 & .005 & Llama-4 & .789 & .008 & .001 & .004 & Cost dominates, cheapest wins \\
   & TD & Gemini-3 & .709 & .144 & .006 & GPT-5.2 & .684 & .152 & .005 & .001 & $M$ weight still decisive \\
   & RC & Mistral & .462 & .000 & .004 & Llama-4 & .501 & .009 & .004 & .000 & Cost dominates, cheapest wins \\
   & RD & Mistral & .545 & .000 & .005 & Llama-4 & .523 & .009 & .004 & .001 & Cost dominates, cheapest wins \\
   & ES & Mistral & 1.00 & .000 & .004 & Llama-4 & 1.00 & .009 & $-$.002 & .006 & Cost dominates, cheapest wins \\
\bottomrule
\multicolumn{12}{l}{\begin{minipage}{0.95\textwidth}
\vspace{1mm}
\footnotesize TC=Ticket Classification, KB=Knowledge Base Search, TD=Technical Diagnosis, RC=Refund Calculation, RD=Response Drafting, ES=Escalation Summary. Gemini-3 = Gemini-3-Pro, Claude-O = Claude-Opus-4.5, Mistral = Mistral-Small-3.1, Llama-4 = Llama-4-Maverick.
\end{minipage}}
\end{tabular}
\end{table*}

\newpage
\subsection{Results from Budget-Based Routing}
\label{apdx:budget_routing}
Table~\ref{tab:budget_routing} demonstrates budget-based routing across three budget levels, assuming $B$ is allocated per 1{,}000 pipeline runs. Assignments are recovered via the dynamic programming algorithm in Section \ref{section:routing_algorithms}, which maximizes total quality-weighted match subject to the absolute cost constraint from Section \ref{section:cost_modeling}, via token estimates from Table~\ref{tab:subtask_profiles}.

\begin{table*}[ht]
\small
\centering
\caption{Routing assignments and system-generated explanations across budget constraints (per 1{,}000 pipeline runs).}
\label{tab:budget_routing}
\setlength{\tabcolsep}{4pt}
\begin{tabular}{c|l|p{10cm}}
\toprule
\textbf{Budget} & \textbf{Assignments} & \textbf{Trace-driven Explanation of Routing Decisions} \\
\midrule
\textbf{\$5.00} & 
\begin{tabular}[t]{@{}l@{}}
TC$\to$Mistral-Small-3.1 \\
KB$\to$Mistral-Small-3.1 \\
TD$\to$Llama-4-Maverick \\
RC$\to$Llama-4-Maverick \\
RD$\to$Mistral-Small-3.1 \\
ES$\to$Mistral-Small-3.1
\end{tabular}
& To maximize quality within the budget, the engine routed lower-complexity tasks like Ticket Classification and Escalation Summary to Mistral, as its summarization and instruction-following capabilities provided the best value for standard requirements. For the highly sensitive Technical Diagnosis and Refund Calculation tasks, the system upgraded to Llama-4-Maverick to leverage its stronger logical and mathematical reasoning. Because top-tier frontier models like GPT-5.2 or Gemini-3-Pro were cost-prohibitive for the entire six-task sequence, these selections represent the most effective performance-to-cost tradeoff available under your constraints.\\
\midrule
\textbf{\$50.00} & 
\begin{tabular}[t]{@{}l@{}}
TC$\to$Mistral-Small-3.1 \\
KB$\to$\colorbox{green!15}{Gemini-3-Pro} \\
TD$\to$\colorbox{green!15}{Gemini-3-Pro} \\
RC$\to$\colorbox{green!15}{Gemini-3-Pro} \\
RD$\to$\colorbox{green!15}{Gemini-3-Pro} \\
ES$\to$Mistral-Small-3.1
\end{tabular}
& We prioritized Gemini for high-stakes tasks like Technical Diagnosis, Refund Calculation, and Response Drafting because these required the advanced logical reasoning, mathematical accuracy, and writing quality necessary to meet your high quality-sensitivity targets. To stay within budget, we routed simpler tasks like Ticket Classification and Escalation Summary to the more economical Mistral, leveraging its efficiency for basic summarization and instruction following where precision was less critical. While elite models like Claude Opus were evaluated, they were excluded to prevent a budget overrun, leaving Gemini as the optimal balance for providing necessary tool-use and complex reasoning capabilities. This strategy ensures that your most complex requirements receive top-tier performance while maintaining overall cost-effectiveness for routine operations.\\

\midrule
\textbf{\$100.00} & 
\begin{tabular}[t]{@{}l@{}}
TC$\to$Mistral-Small-3.1 \\
KB$\to$Gemini-3-Pro \\
TD$\to$\colorbox{green!15}{Claude-Opus-4.5} \\
RC$\to$Gemini-3-Pro \\
RD$\to$Gemini-3-Pro \\
ES$\to$Mistral-Small-3.1
\end{tabular}
& We prioritized the most resource-intensive task, Technical Diagnosis, by selecting Claude-Opus to leverage its superior logical reasoning and tool-use capabilities where quality sensitivity was at its maximum. To offset this premium cost, we routed low-complexity tasks like Ticket Classification and Escalation Summary to Mistral, which provides sufficient summarization and extraction skills at a fraction of the price. The remaining high-sensitivity tasks, such as Refund Calculation and Response Drafting, were assigned to Gemini to ensure high-tier mathematical and writing precision without exceeding the total budget. This strategy successfully balances peak performance for critical reasoning steps with aggressive cost-saving on tasks where the user indicated lower quality requirements.\\
\bottomrule

\multicolumn{3}{l}{\begin{minipage}{\textwidth}
\vspace{0.5mm}
\footnotesize TC=Ticket Classification, KB=Knowledge Base Search, TD=Technical Diagnosis, RC=Refund Calculation, RD=Response Drafting, ES=Escalation Summary.
\end{minipage}}
\end{tabular}
\vspace{-4mm}
\end{table*}

\paragraph{Allocation Behavior Across Budgets.}
At \textbf{\$5}, the optimizer is severely constrained and concentrates its limited premium budget on the two tasks of highest importance, Technical Diagnosis ($q=1.0$) and Refund Calculation ($q=0.95$). Those tasks are routed to Llama-4 Maverick for additional reasoning capabilities, while all other tasks route to Mistral to stay below budget. At this limited budget, the router cannot afford any of the more expensive models. 

At \textbf{\$50}, the optimizer can afford Gemini-3 Pro for most tasks. Knowledge Base Search, Technical Diagnosis, Refund Calculation, and Response Drafting all upgrade from Mistral or Llama to Gemini, whose advanced logical reasoning, mathematical precision, and writing quality justify the spend given the tasks' quality sensitivities. Ticket Classification and Escalation Summary remain on Mistral---their task complexity and quality sensitivity are low enough that upgrading models does not add enough quality to justify the cost, as Mistral is able to do a good enough job. While Claude Opus was evaluated, its cost would have caused a budget overrun; Gemini provides the optimal balance.

At \textbf{\$100}, the only change from \$50 is that Technical Diagnosis upgrades to Claude Opus 4.5, whose superior logical reasoning and tool-use capabilities yield a meaningful match score improvement for the pipeline's most complex task. The remaining assignments are unchanged, as no other upgrade adds as much quality-per-dollar, and this assignment uses practically all of the budget.

\end{document}